\documentclass{article} 

\usepackage{iclr2026_conference,times}
\usepackage{amsmath,amssymb,amsthm}
\usepackage{booktabs}
\usepackage{array}
\usepackage{multirow}
\usepackage{graphicx}
\usepackage{xcolor}
\usepackage{tikz}
\usetikzlibrary{arrows.meta,positioning}
\usepackage{hyperref}
\usepackage{url}

\graphicspath{{figs/}}
\iclrfinalcopy

\newtheorem{proposition}{Proposition}
\title{One Axis, No Brake: Self-Knowledge Limits the Filtering of
Harmful Peer Conformity in LLMs}
\author{Yibo Hu \\ Illinois Institute of Technology \\ \texttt{yhu89@illinoistech.edu}}

\begin{document}
\maketitle
\lhead{}

\begin{abstract}
Multi-agent LLM systems are expected to be more reliable because agents can catch
each other's mistakes.
But peer pressure cuts both ways: the same correction that
fixes a wrong answer can overturn a right one. The tempting safeguard is a brake that
keeps the beneficial revisions and blocks the harmful ones. We show this brake is hard
to build, for a simple reason: a revision is harmful exactly when the original answer
was right, so deciding whether to block it is the same as knowing whether the model was
already correct. This turns the open-ended hunt for a brake into one measurable
quantity, the model's self-knowledge: any brake built from a deploy-time signal is a
correctness probe in disguise, and self-knowledge is far from perfect
(AUROC $\approx 0.64$--$0.89$ across six model families). We call this ceiling the
\emph{wall}. Even white-box steering of the model's own correctness direction does not
breach it: it changes how often the model revises, but harmful and beneficial revisions
move together. At population scale the wall becomes the \emph{cliff}: when most agents
start wrong, debate amplifies the shared mistake into a confident, wrong consensus. In
our multiple-choice societies, more agents, more model diversity, and a stronger member
do not fix it. What helps is adding information before the revision, not filtering after
it. Local agreement is not global correctness.\footnote{Code and data: \url{https://github.com/yibo-hu-lab/wall-and-cliff}}
\end{abstract}

\section{Introduction}
\label{sec:intro}

Multi-agent LLM systems are increasingly deployed on the premise that conferring
makes them more reliable: let several agents debate, review one another, or vote, and
the group should catch mistakes a single model would miss. The premise has support
(debate and self-consistency can raise accuracy \citep{du2023debate,wang2023selfconsistency,li2024moreagents}),
and it now underwrites production pipelines for reasoning, code, and decision support.

But the same exchange can backfire: it can pull a model off a right answer as easily as
it fixes a wrong one. Picture a model that answers a question
correctly, sees three peers disagree, and revises to match them: the ``correction''
has made it worse. Call a revision \emph{harmful} when it abandons a correct answer
and \emph{beneficial} when it fixes a wrong one. A system that wanted the benefit
without the harm would keep the beneficial revisions and block the harmful ones.

A large literature documents that LLMs are swayed by users, peers, and majority
pressure \citep{sharma2023sycophancy,perez2022discovering,cho2025herd}, and that
multi-agent debate can fail or even degrade accuracy
\citep{wuli2025majority,demarzo2026misalignment}. Most of it asks one of two things:
\emph{whether} models are influenced, and whether conferring improves \emph{aggregate}
accuracy. We ask the question a deployed system faces one step later: once revisions
happen, can the system tell its harmful revisions from its beneficial ones and filter
accordingly?

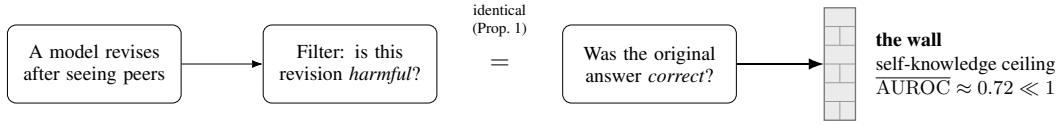
\begin{figure}[t]
\centering
\resizebox{\linewidth}{!}{%
\begin{tikzpicture}[
  every node/.style={font=\footnotesize},
  box/.style={draw,rounded corners,align=center,inner sep=4pt,
              text width=2.5cm,minimum height=1.25cm}]
\node[box] (rev) {A model revises after seeing peers};
\node[box,right=13mm of rev] (filt) {Filter: is this revision \emph{harmful}?};
\node[box,right=20mm of filt] (corr) {Was the original answer \emph{correct}?};
\draw[-{Latex}] (rev)--(filt);
\draw[draw=none] (filt)--(corr)
  node[midway,font=\large]{$=$}
  node[midway,yshift=7mm,font=\scriptsize,align=center]{identical\\(Prop.~\ref{prop:reduction})};
\draw[-{Latex},line width=0.9pt] (corr.east)--([xshift=14mm]corr.east);
\fill[black!8]            ([xshift=14mm,yshift=-9mm]corr.east) rectangle ([xshift=19mm,yshift=9mm]corr.east);
\draw[black!55,line width=0.6pt] ([xshift=14mm,yshift=-9mm]corr.east) rectangle ([xshift=19mm,yshift=9mm]corr.east);
\draw[black!40,line width=0.4pt] ([xshift=14mm,yshift=-6mm]corr.east)--([xshift=19mm,yshift=-6mm]corr.east);
\draw[black!40,line width=0.4pt] ([xshift=14mm,yshift=-3mm]corr.east)--([xshift=19mm,yshift=-3mm]corr.east);
\draw[black!40,line width=0.4pt] ([xshift=14mm,yshift=0mm]corr.east)--([xshift=19mm,yshift=0mm]corr.east);
\draw[black!40,line width=0.4pt] ([xshift=14mm,yshift=3mm]corr.east)--([xshift=19mm,yshift=3mm]corr.east);
\draw[black!40,line width=0.4pt] ([xshift=14mm,yshift=6mm]corr.east)--([xshift=19mm,yshift=6mm]corr.east);
\draw[black!40,line width=0.4pt] ([xshift=16.5mm,yshift=-9mm]corr.east)--([xshift=16.5mm,yshift=-6mm]corr.east);
\draw[black!40,line width=0.4pt] ([xshift=16.5mm,yshift=-3mm]corr.east)--([xshift=16.5mm,yshift=0mm]corr.east);
\draw[black!40,line width=0.4pt] ([xshift=16.5mm,yshift=3mm]corr.east)--([xshift=16.5mm,yshift=6mm]corr.east);
\node[align=left,anchor=west] at ([xshift=21mm]corr.east)
  {\textbf{the wall}\\[1pt] self-knowledge ceiling\\[1pt] $\overline{\mathrm{AUROC}}\approx 0.72 \ll 1$};
\end{tikzpicture}}
\caption{The reduction, and the wall it hits. Blocking a harmful peer revision means
knowing the original answer was right, but on the set of revisions that is the
\emph{same} question (Proposition~\ref{prop:reduction}). Answering it is the model's
self-knowledge, so every post-hoc filter runs into one ceiling, the \emph{wall}:
separability is imperfect (mean AUROC $\approx0.72$; per-family range
$0.64$--$0.89$, \S\ref{sec:ceiling}).}
\label{fig:reduction}
\end{figure}

The answer is unexpectedly simple, and it is a constraint rather than a method. On the
set of revisions, a revision is harmful exactly when the original answer was correct.
So asking ``is this revision harmful?'' is, term for term, asking ``was I right to
begin with?'' Any filter that runs on signals a model exposes at deploy time (its
verbalized confidence, its logprobs, its hidden activations) is therefore a
correctness probe (Figure~\ref{fig:reduction}). It can be no
better at catching harmful revisions than the model is at knowing when it was right.
And self-knowledge is far from perfect: across the six model families we test, the best
deploy-time read of a model's own correctness reaches AUROC only $\approx 0.64$--$0.89$,
the same imperfect regime prior work reports for internal truth signals
\citep{kadavath2022know,azaria2023internal}. We call this ceiling the \emph{wall}.

The wall is not the weakness of one signal. We hit it with three: a coverage-calibrated
confidence gate, an engineered hidden-state probe, and (the sharpest test) causal
activation steering along the model's own correctness direction. Even with full
white-box access, steering buys no selective brake: turning the knob changes how
\emph{often} a model revises, but harmful and beneficial revisions rise and fall
together. Steering moves revision \emph{propensity}, not revision \emph{valence}.

The wall has a consequence the moment agents revise \emph{each other}. \emph{The cliff
is the wall at population scale, not a separate phenomenon.}
Because no agent can reliably brake its own unwarranted revision, a group whose members
err in similar ways has no internal channel to recover, and its fate is set by where it
starts. Groups that begin mostly correct converge to the truth; groups that begin
mostly wrong (the hard items) amplify the shared mistake into a confident, wrong
consensus. The collapse is difficulty-gated, and it resists the rescues one reaches for
first: more agents, more model diversity, a stronger member, even a correct dissenting
minority.

The design lesson follows directly. Multi-agent debate is a useful \emph{exploration}
mechanism (it surfaces candidates and covers ground a lone agent would not), but it is
not a self-certifying \emph{safety} mechanism: it cannot filter its own errors on the
problems where filtering matters most. What helps is not a better post-hoc filter but
adding information before or around the revision: an external verifier or retrieved
evidence, decorrelated peers, or a pre-committed warrant for what would justify a
change. In one line: \emph{post-hoc filtering of harmful peer revision is self-knowledge
in disguise, and because self-knowledge is imperfect, multi-agent LLM systems can lock
into confident, wrong consensus on hard items.}

\paragraph{Contributions.}
\begin{enumerate}
\item \textbf{Filtering reduces to self-knowledge.} On the set of revisions, telling
  a harmful revision from a beneficial one is identical to reading one's own initial
  correctness (Proposition~\ref{prop:reduction}); the empirical ceiling is far from
  perfect ($\approx 0.64$--$0.89$) and holds across families, scale, and reasoning.
\item \textbf{White-box access does not create a selective brake.} Causal steering of
  the correctness direction, both linear and sparse-autoencoder (SAE) feature, changes how often a model revises but not which revisions
  ($|\Delta H-\Delta B|<0.05$ in the deploy-relevant range at $\le$8B; the arms stay together through 32B).
\item \textbf{The same limit creates a population cliff.} Groups that start mostly
  wrong lock into confident-wrong consensus, and in the regimes we test more agents,
  diversity, and a stronger member do not help.
\item \textbf{Useful interventions add information before revision.} Decorrelating the
  peer set measurably cuts harmful revisions ($\approx2.2\times$); a post-hoc filter, by
  the reduction, cannot. (A pre-committed warrant is a further, still-preliminary lever; \S\ref{sec:levers}.)
\end{enumerate}

\section{Setup}
\label{sec:setup}

\paragraph{Protocol.} A model first answers a multiple-choice question on its own
(round~1). It is then shown a short summary of how a panel of peers answered
(reported as ``$k$ chose (X)'', with or without the peers' average confidence, and
constructed to disagree with it) and answers again (round~2). The summary carries only
the social signal: how many peers disagree and how confident they are. It adds no
evidence or argument. We vary this channel to isolate the effect of social pressure
itself, not persuasion by content. The \emph{revision
set} $F$ is the items on which the round-2 answer differs from round-1; we condition
everything that follows on $F$, since only a revision can be harmful or beneficial.
We use four open models (Qwen2.5-7B, Llama-3.1-8B, Mistral-7B, gemma-2-9B) on
ARC-Challenge and TruthfulQA, plus a Qwen2.5 scale ladder (1.5B--72B) and Qwen2.5-3B,
with targeted grids for the probe, steering, and
population studies (Appendix~\ref{app:details}); self-knowledge AUROCs use $n{=}200$
items per cell.

\paragraph{What we measure.} Let $\textit{correct}_0$ be round-1 correctness. On a
revision, define its valence $V = 1-\textit{correct}_0$, so $V{=}0$ marks a
\textbf{harmful} revision (the model abandoned a correct answer) and $V{=}1$ a
\textbf{beneficial} one (it fixed a wrong answer). A \emph{deploy-time signal} $S$ is
anything available without ground truth: verbalized confidence, logprob margin or
entropy, or hidden activations. A \emph{brake} is a score $g(S)$ that ranks harmful
revisions above beneficial ones on $F$ well enough to act on. Equivalently, it detects
the harmful indicator $H \equiv \textit{correct}_0$ (so $H{=}1$ on harmful revisions),
which we take as the positive class. We grade any candidate brake by how well it
separates the two classes, measured by AUROC (area under the ROC curve; $1$ =
perfectly separable, $0.5$ = chance). AUROC scores separability and is
unchanged up to orientation ($\mathrm{AUROC}\!\leftrightarrow\!1-\mathrm{AUROC}$) by
which class is called positive. We write $\mathrm{AUROC}^{*}_{Y}(S\mid F)$ for the
largest AUROC achievable for a target label $Y$ by any score derived from $S$;
empirically we approximate it by the strongest scoring rule we can fit, so the
reported numbers are lower bounds.

\section{The Wall: Filtering Is Bounded by Self-Knowledge}
\label{sec:wall}

\subsection{Why filtering needs self-knowledge}
\label{sec:reduction}

A harmful revision is, by definition, a revision away from a correct answer, and a
beneficial one a revision away from a wrong answer. So on the revisions, labeling a
revision harmful is exactly labeling the initial answer correct.

\begin{proposition}[Reduction]
\label{prop:reduction}
On the revision set $F$, valence is fixed by initial correctness, $V = 1-\textit{correct}_0$.
Telling a harmful revision from a beneficial one with a score $g(S)$ is therefore the same
discrimination as reading $\textit{correct}_0$ from $g(S)$, up to the orientation of the
score, so $\sup_g \mathrm{AUROC}_V(g(S)\mid F) = \mathrm{AUROC}^{*}_{\textit{correct}_0}(S\mid F)$:
separating harmful from beneficial revisions is the same problem as reading round-1
correctness, and is bounded by the model's self-knowledge in $S$.
\end{proposition}

The reduction is elementary, and that is its force: it collapses the open-ended
search for a deploy-time brake (across confidence rules, hidden-state probes, and
activation edits) into a single measurable quantity: how well the model knows its own
correctness. The bound
needs only that correctness be checkable, not multiple choice specifically; whether a
model's deploy-time signals stay below it on other checkable tasks is then an empirical
question. What
remains is empirical: how high that ceiling sits, and whether any signal reaches it.
(Selection slack and a data-processing argument for causal interventions are in
Appendix~\ref{app:proof}.)

A concrete case from our runs makes the identity tangible. Asked why an ice cube
melts in a hand, Qwen2.5-7B first answers correctly, ``heat moves from her hand to the
ice cube''; shown three peers who unanimously pick the reverse, it revises to their
answer. To block this one harmful revision, a filter would have to recognize that the
original answer was the correct one, that heat flows from warm to cold, which is
precisely the correctness judgment of Proposition~\ref{prop:reduction}. A deploy-time
signal that could reliably flag the flip would be one that already knew the model was
right, so there is no shortcut around self-knowledge.

\subsection{Every signal reaches the same ceiling}
\label{sec:ceiling}

\begin{figure}[t]
\centering
\includegraphics[width=0.8\linewidth]{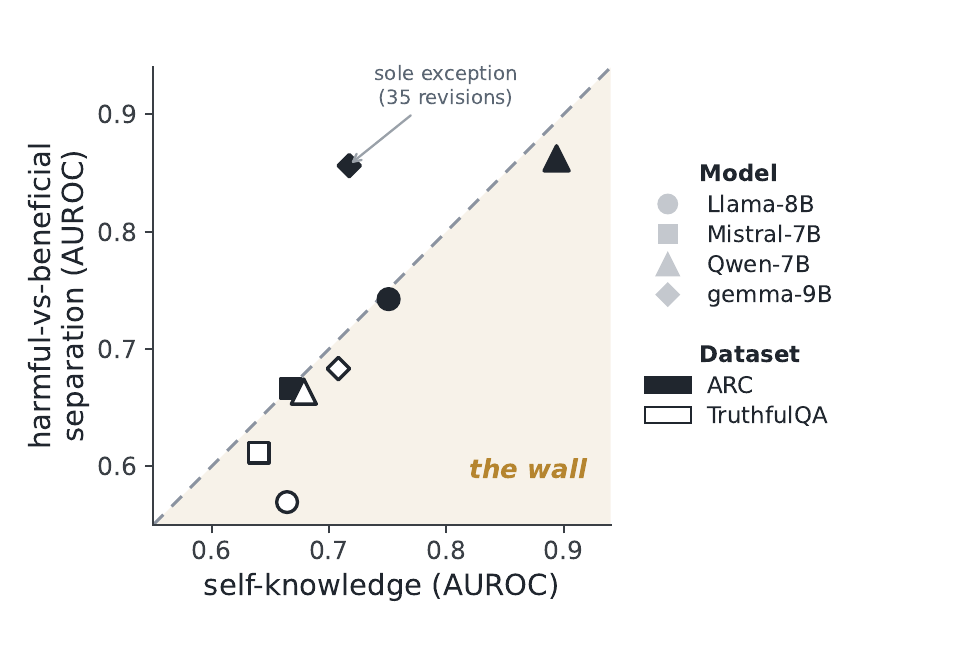}
\caption{The wall. For each model and dataset, separating harmful from beneficial
revisions (\emph{y}) is no easier than reading one's own correctness (\emph{x}): all
points but one (gemma/ARC, a 35-revision cell) lie on or below $y{=}x$.}
\label{fig:ceiling}
\end{figure}

Figure~\ref{fig:ceiling} shows the wall directly: across models and datasets,
telling a harmful revision from a beneficial one is no easier than reading one's
own correctness: seven of eight cells lie on or below the $y{=}x$ line, and the one
exception (gemma/ARC) is a 35-revision cell whose gap CI spans $0$ (per-cell values
in Appendix~\ref{app:ceiling}).

Three signals confirm it. A coverage-calibrated confidence gate catches $\approx 95\%$
of harmful revisions only by blocking most beneficial ones, and does no better than
plain confidence, at any operating point, not just in AUROC
(Figure~\ref{fig:gate}; per-threshold curve and net-accuracy detail in Appendix~\ref{app:ceiling}). A hidden-state probe, even when engineered (late-layer
pressure-induced deltas), reaches AUROC $0.81$--$0.83$ and does not exceed the
$\approx 0.75$--$0.83$ self-knowledge band; an answer-content control confirms it
reads genuine correctness rather than the answer change. A classifier trained on the
model's own free-text \emph{rationale} for the revision does worse still. It separates
harmful from beneficial revisions at chance (AUROC $0.46$):
the model's stated reasoning carries no usable signal about whether its own revision was
harmful. Two further checks stay under the same ceiling. A logistic-regression ensemble
of the three deploy-time signals (verbalized confidence, logprob confidence, logprob
entropy), scored by out-of-fold cross-validation, does not beat the best single one
(mean $\Delta=-0.05$ AUROC).
Sampling-based self-consistency uncertainty, a signal
class outside the single-pass argument, reaches only AUROC $\approx0.71$ (Llama-3.1-8B)
and does not improve as the sample count grows to $24$ (Appendix~\ref{app:ceiling}). The
third signal, causal steering, is the sharpest test
and the subject of \S\ref{sec:steering}.

\begin{figure}[t]
\centering
\includegraphics[width=0.5\linewidth]{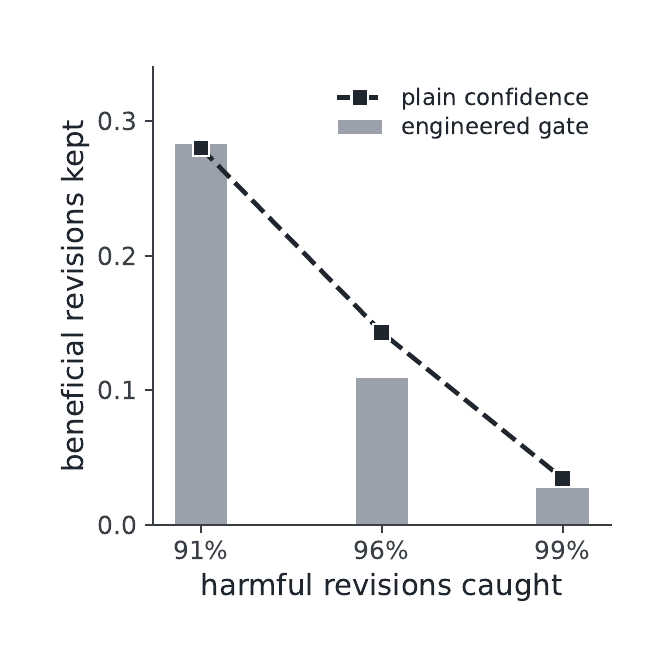}
\caption{No selective brake. Beneficial revisions kept vs.\ harmful revisions caught,
at three operating points: the engineered gate (grey bars) never rises to plain
confidence (dashed line), so catching more harm always costs more benefit.}
\label{fig:gate}
\end{figure}

\subsection{White-box steering gives no clean brake}
\label{sec:steering}

The sharpest test is causal. We steer activations along the model's own correctness
direction (the most direct way to make a model ``trust itself'' more or less) and
ask whether that pulls harmful and beneficial revisions apart. It does not. In the
deploy-relevant range, steering changes how often the model revises but not
\emph{which} revisions: the harmful and beneficial arms move together, and a
pre-registered equivalence test confirms a gap below $\pm0.05$
(Figure~\ref{fig:steering}; ranges and full grid in Appendix~\ref{app:steering}). Pushing the knob harder does not recover a brake:
extreme steering cuts harmful revisions only by cutting beneficial ones too. Steering moves revision
\emph{propensity}, not revision \emph{valence}. The null is not an artifact of the linear
direction: an independent sparse-autoencoder steering direction gives the same null on
Llama-3.1-8B, on both ARC and TruthfulQA (Appendix~\ref{app:steering}). White-box access, linear or
SAE-feature, buys no selective brake. The same null holds at 32B on ARC: the two arms
stay together in the deploy-relevant range (max $|\Delta H-\Delta B|=0.07$ for $|\alpha|\le1$, accuracy flat; Appendix~\ref{app:steering}). The beneficial arm is small there, so we read 32B as consistent with the null rather than as a clean equivalence test.

\begin{figure}[t]
\centering
\includegraphics[width=\linewidth]{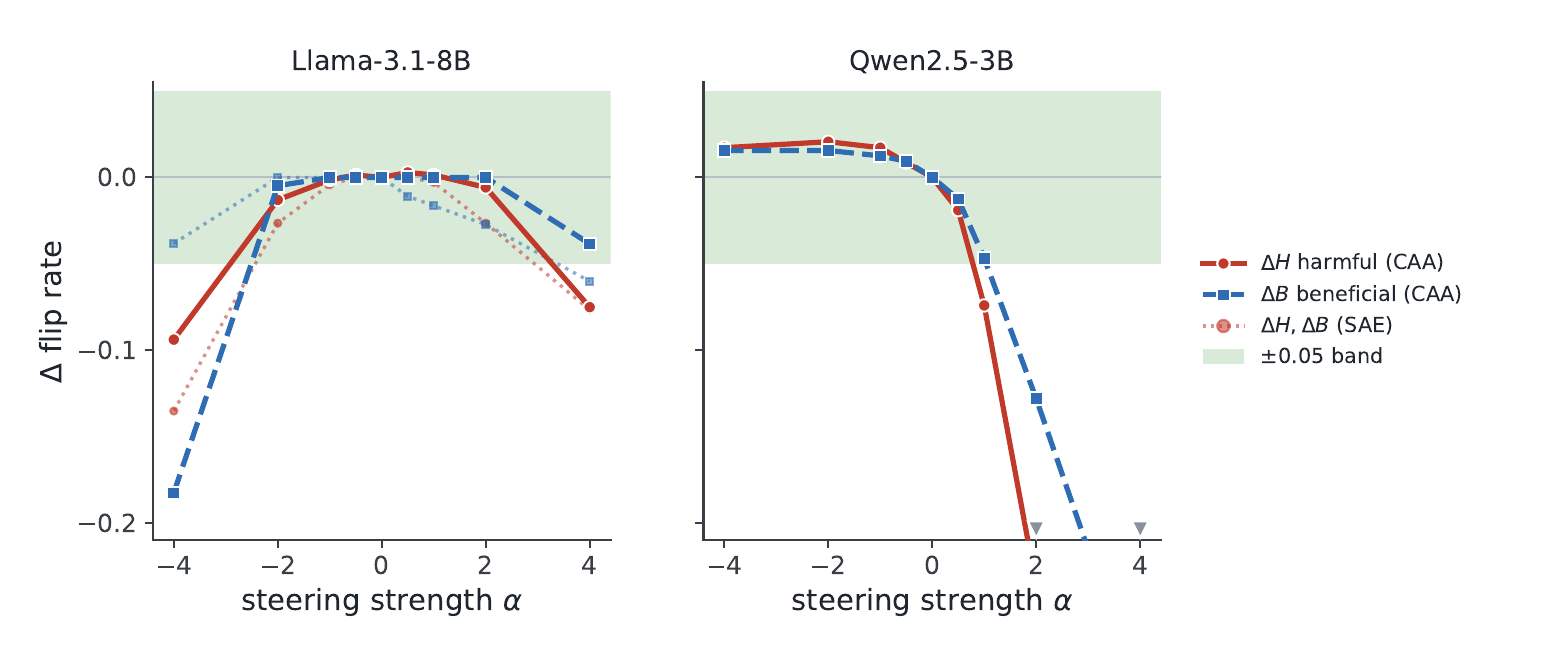}
\caption{Steering moves how \emph{often} a model revises, not \emph{which}. As $\alpha$
varies, the harmful ($\Delta H$) and beneficial ($\Delta B$) arms rise and fall
\emph{together}: inside the $\pm0.05$ band in range, collapsing together at extreme
$\alpha$, so white-box access buys no selective brake. Dotted lines (Llama) use an
SAE-feature direction (Goodfire L19) with the same null; grey triangles mark $\alpha$
beyond the plotted range (full values in Appendix~\ref{app:steering}).}
\label{fig:steering}
\end{figure}

\subsection{The wall is structural}
\label{sec:structural}

The ceiling is not an artifact of scale, family, or quantization
(Table~\ref{tab:ceiling}). It stays well below $1$ for all six families, including
two full-precision held-out models never tuned on our stack. It does not rise across
a 1.5B$\to$72B scale ladder (plateau $0.73$--$0.83$; at 72B, ARC $0.81$ / TQA $0.67$,
three seeds each). Test-time reasoning does not lift it either (R1-distilled models
reach $0.64$--$0.72$ on ARC). Across the six
families the self-knowledge AUROC spans $0.64$ (Mistral/TQA) to $0.89$ (Qwen-7B/ARC).

\begin{table}[t]
\centering\small
\caption{The wall is structural: self-knowledge AUROC (the robust quantity,
$n{=}200$ per cell) stays $\ll 1$ across precision, scale, and reasoning.}
\label{tab:ceiling}
\setlength{\tabcolsep}{5pt}
\begin{tabular}{llcc}
\toprule
Axis & Cell & self-know.\ AUROC & note \\
\midrule
\multirow{2}{*}{held-out, bf16}
 & OLMo-2-7B (ARC / TQA)    & $0.79$ / $0.66$ & adequate \\
 & Phi-3.5-mini (ARC / TQA) & $0.77$ / $0.85$ & few revisions \\
\midrule
\multirow{5}{*}{scale ladder}
 & 1.5B & $\approx$ chance & \\
 & 7B   & $0.78$ & \\
 & 14B  & $0.73$--$0.77$ & \\
 & 32B  & $0.76$--$0.83$ & plateau \\
 & 72B  & $0.81$ & plateau (TQA $0.67$) \\
\midrule
\multirow{3}{*}{R1-distill}
 & R1-Qwen-7B / ARC  & $0.64$ & confound-free \\
 & R1-Llama-8B / ARC & $0.72$ & confound-free \\
 & R1-Qwen-7B / TQA  & $0.73$ & confound-free \\
\bottomrule
\end{tabular}
\end{table}

\paragraph{Beyond multiple choice.} The wall is not an artifact of the multiple-choice
format. On two exact-match-checkable free-response tasks, GSM8K and the harder MATH,
self-knowledge stays imperfect (best-signal AUROC $0.58$--$0.85$ across two models each,
always $<1$; Appendix~\ref{app:transfer}).

\section{The Cliff: Populations Lock Into Confident-Wrong Consensus}
\label{sec:cliff}

\begin{figure}[t]
\centering
\includegraphics[width=0.5\linewidth]{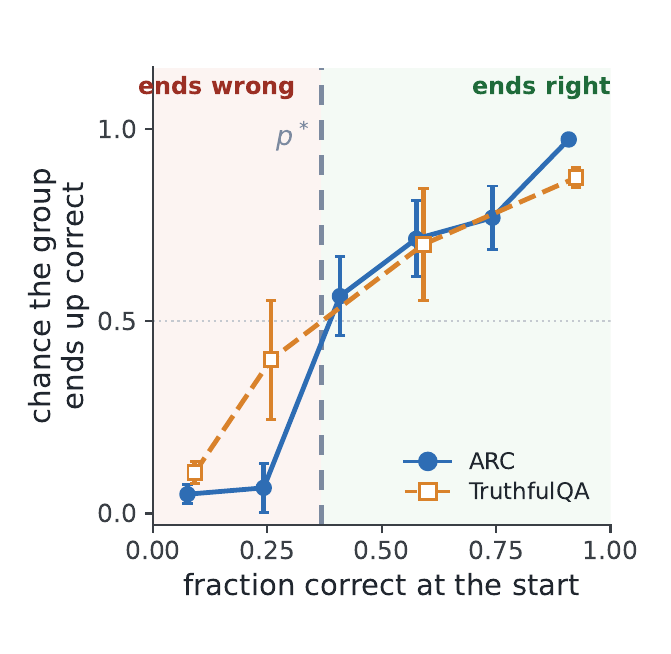}
\caption{The cliff. Chance the final consensus is correct vs.\ the fraction of agents
correct at the start, pooled over society runs. Groups that start mostly wrong lock into
a confident-wrong consensus (``ends wrong''); those that start mostly right converge to
the truth (``ends right''). The dashed line $p^*$ is a mean-field separatrix predicted
from measured per-agent quantities (no fit to these runs) and falls near the crossover.}
\label{fig:basin}
\end{figure}

The cliff is not a separate phenomenon: it is what the wall looks like once agents
revise each other. Because an agent cannot reliably filter its own unwarranted
revision, a group whose members err in similar ways has no internal way to
recover, so the outcome depends sharply on where the group starts
(Figure~\ref{fig:basin}). Groups that start mostly correct converge to truth; groups
that start mostly wrong (the hard items) amplify the shared mistake into a
confident, wrong consensus (final-consensus accuracy $\approx 0.00$--$0.32$ vs.\
$0.71$--$1.00$ on TruthfulQA; $0.00$--$0.36$ vs.\ $0.94$--$1.00$ on ARC). The gap is
significant in every run, and confident-wrong lock-in covers $25$--$55\%$ of hard
items on TruthfulQA and $5$--$18\%$ on the easier ARC.

The cliff's location is not arbitrary. A simple model, evaluated at each model's
measured calibration with no fit to these runs, places the separatrix at
$p^{*}\in[0.357,0.391]$ across the four models, which falls at the observed crossover
(Figure~\ref{fig:basin}, dashed line $p^*$; Appendix~\ref{app:meanfield}), so the
population transition sits where the measured per-agent quantities predict, without any fit to the runs.

The wall is what leaves the cliff un-braked. A hypothetical selective brake (a
correctness detector of quality $q=\mathrm{AUROC}$ that blocks harmful revisions at its
best operating point) must reach $q\gtrsim0.96$ to avert the collapse, whereas a brake
at the measured self-knowledge ceiling ($q\le0.89$) leaves confident-wrong lock-in on
the hardest items at $0.41$--$0.98$ (Appendix~\ref{app:meanfield}).
The cliff is the population-scale consequence of the same bounded self-knowledge.

\subsection{The driver is majority dynamics, not confidence}
\label{sec:cliff-mechanism}

Hiding vs.\ showing peer confidence makes no difference to the lock-in, on both
datasets; \citet{wuli2025majority} report the same. So the driver is majority
dynamics on shared misconceptions, not confidence weighting, and we treat the
mean-field model as illustrative rather than a validated mechanism.

\subsection{The collapse resists the obvious fixes}
\label{sec:cliff-robustness}

None of the natural rescues works (Table~\ref{tab:robust}). The lock-in is flat from
$N{=}6$ to $N{=}24$ agents. A heterogeneous society scores below its best member
in all three seeds ($0.60$--$0.66<0.71$--$0.77$ on TruthfulQA; corroborated on ARC,
Appendix~\ref{app:robust}). A correct minority is outvoted $71$--$100\%$ of the time on
the qualifying items.

\begin{table}[t]
\centering\small
\caption{Every obvious rescue fails in the tested setting. Each row is a fix a
reader might propose; none recovers the group on hard items (TruthfulQA).}
\label{tab:robust}
\setlength{\tabcolsep}{6pt}
\begin{tabular}{lll}
\toprule
``But what if you\ldots'' & What we observe & Verdict \\
\midrule
add more agents? & gap, lock-in, convergence flat ($N{=}6\!\to\!24$) & no help \\
use diverse models? & society $<$ best, 3/3 seeds ($0.60$--$0.66$ vs $0.71$--$0.77$) & falls below best \\
let a correct minority rescue it? & minority outvoted $71$--$100\%$ ($n{\approx}5$--$21$) & outvoted \\
\bottomrule
\end{tabular}
\end{table}

\section{What Helps: Acting Before the Revision}
\label{sec:levers}

Since no post-hoc gate separates harmful from beneficial revisions, the useful
levers act before or around it. \textbf{Decorrelating the peer set} helps: a
unanimous wrong panel multiplies the harmful-revision rate $\approx 2.2\times$ over a
mixed panel ($2.35\times$ on ARC, $1.99\times$ on TruthfulQA), so not manufacturing a
fake consensus is a deployable rule.

\textbf{A pre-registered warrant} (committing
in advance to what would justify a change) is a preliminary transparency lever: in
an ARC audit it exposes a subset of harmful revisions whose stated reasons depart
from that pre-commitment, a signal post-hoc text classifiers miss
(Appendix~\ref{app:warrant}). Both work with the wall rather than against it.

\section{Related Work}
\label{sec:related}

\paragraph{Conformity, sycophancy, and peer influence.}
LLMs bend to social pressure. Trained with human feedback, assistants come to favor
user-matching responses over truthful ones \citep{sharma2023sycophancy}, a tendency
measurable at scale \citep{perez2022discovering} that splits into accuracy-improving
and accuracy-harming revisions \citep{fanous2025syceval}; internally it is a largely
linear, sometimes deep-layer override of intact knowledge
\citep{sycophancy_override2025}, and ``not one thing'': genuine and sycophantic
agreement are causally separable \citep{sycophancy_notone2025}. The same bending
appears between agents, as herd behavior and inter-agent sycophancy
\citep{cho2025herd,yao2025peacemaker}, and at the single-agent level a wrong consensus
flips correct answers more readily than a correct consensus repairs wrong ones
\citep{qu2026easier}. This work establishes \emph{that} models conform and
characterizes sycophancy behaviorally and internally. We take the step a deployed
system needs next: whether the harmful half of that conformity can be filtered out
\emph{after} the revision, and show that doing so reduces to reading one's own
correctness.

\paragraph{Multi-agent debate and collective decision-making.}
Debate, self-consistency, and voting are widely reported to improve accuracy
\citep{du2023debate,wang2023selfconsistency,li2024moreagents}, though the gains are
contested: a single well-prompted agent can match a discussion
\citep{wang2024rethinking}, and debate can entrench rather than correct an error
\citep{wuli2025majority}. At population scale, individually aligned agents settle into
a stable, collectively misaligned consensus past a tipping point
\citep{demarzo2026misalignment,committed2025minority}, and system-level taxonomies
trace such failures to interaction structure rather than any single bad agent
\citep{cemri2025mast,bisconti2025esrh}. Some conditions recover debate: answer
diversity with calibrated confidence updates \citep{zhu2026diversity}, or human-
inclusive rather than pure-LLM groups \citep{sheffer2025confident}. These results
establish \emph{that} collectives help or fail. We supply the micro-to-macro
mechanism: a per-agent self-knowledge ceiling becomes a group-level, difficulty-gated
confident-wrong cliff that survives the rescues: more agents, member heterogeneity
(distinct from the answer-initialization diversity of \citet{zhu2026diversity}), a
stronger member, and a correct minority.

\paragraph{Calibration, self-knowledge, and uncertainty signals.}
A line of work asks whether models know what they know. Internal $P(\text{True})$
signals exist but degrade out of distribution \citep{kadavath2022know}; hidden states
carry a truth signal readable at $71$--$83\%$ \citep{azaria2023internal}; a probe at the
exact-answer token reads correctness more sharply on some tasks yet is skill-specific
and does not transfer across them \citep{orgad2025llms}; verbalized
confidence is systematically overconfident and worst on hard items
\citep{xiong2024can}; and models cannot reliably self-correct their reasoning without
external feedback \citep{huang2024selfcorrect}. These signals read correctness
imperfectly, and that imperfection is exactly the wall. Because separating harmful
from beneficial revisions \emph{is} reading correctness
(Proposition~\ref{prop:reduction}), every such signal is a candidate brake and every
one inherits the same ceiling; none is an independent post-hoc channel.

\paragraph{Activation steering and internal control.}
Steering activations along a learned direction can change targeted behavior
\citep{rimsky2023caa}, but is often unstable or non-selective: truthfulness steering
needs adaptive intensity and clustered vectors \citep{wang2024act}, multi-attribute
steering needs sparse orthogonal vectors to avoid cross-talk
\citep{nguyen2025matsteer}, directions transfer across models
\citep{wang2025expertsteer}, and a single self-preference direction flips both
legitimate and illegitimate cases \citep{roytburg2025breakingmirror}. Where one
assumed direction conflates two functions, decomposing it can restore selectivity
\citep{zhang2025dbdi,sycophancy_notone2025}. We run the analogous test on the
correctness axis and obtain a negative result: steering the model's own correctness
direction moves revision propensity but not valence ($\Delta H\approx\Delta B$), so
even white-box access yields no selective brake. Post-hoc detectors of
rationale--answer inconsistency \citep{fan2025peerguard} face the same bound on the
revision set unless they add a channel independent of the answer.

\paragraph{Social dynamics and information cascades.}
Confident-wrong consensus has classical roots: DeGroot averaging
\citep{degroot1974consensus}, bounded-confidence opinion dynamics
\citep{hegselmann2002opinion}, and limits on the wisdom of crowds under influence
concentration \citep{golub2010wisdom}, with herd and cascade behavior now ported to
LLM groups \citep{cho2025herd,cau2025lodas}. In human conformity, people bend to a
wrong unanimous majority but a single dissenting ally breaks the spell
\citep{asch1955opinions}, and a committed minority past a critical mass flips a
convention \citep{centola2018experimental,committed2025minority}; adversarial
minorities steer LLM debates the same way, modulated by interaction topology
\citep{kraidia2026adversary,adversarial2025consensus,topology2026}. None of these
carries a correctness or calibration axis, and the human brake is absent: in our LLM
societies the lone correct dissenter is not Asch's liberating ally but is outvoted, so
the cliff is governed by initial accuracy and (mis)calibration, not by influence
topology alone.

\section{Discussion}
\label{sec:discussion}

\paragraph{What the wall does and does not say.} The wall is about \emph{post-hoc}
control: it says a deployed system cannot tell its harmful revisions from its
beneficial ones using a signal it already has, because that judgment is the same as
knowing whether it was right. It does not say conformity is irrational or that
debate is useless. It says debate cannot \emph{manufacture} correctness the
participants do not already hold, and that the natural knob (be more or less
swayed by peers) moves good and bad revisions together.

\paragraph{When do collectives help, then?} Exactly when a member opens an
independent channel to the answer: re-deriving the problem from scratch, calling a
tool or retriever, or checking verifiable evidence. These add information the group
did not have, which is what the wall requires and what a post-hoc text filter cannot
supply. This reframes a common design choice: on hard items, scaling the number or
diversity of agents trades against the cliff, while adding one verifier with a real
channel to ground truth is what changes the outcome.

\paragraph{The wall is highest where it hurts most.} The individual limit and the
population catastrophe are coupled by difficulty. Self-knowledge is not uniform:
verbalized confidence is most overconfident on hard items \citep{xiong2024can}, and
correctness probes are skill-specific rather than general \citep{orgad2025llms}, so
the wall rises exactly on the questions a model is most likely to answer wrong. Those
are the same items on which a group starts mostly wrong and locks in
(\S\ref{sec:cliff}). The post-hoc brake is therefore weakest precisely where the
population is most vulnerable: individual self-doubt and collective lock-in peak on
the same inputs, which is why a fix has to add an outside channel rather than tune the
group's internal dynamics.

\paragraph{Practical reading.} This is why the levers that survive the wall
(\S\ref{sec:levers}) are design and transparency choices rather than detectors: the
only two routes that move the bound are raising calibration, so the wall sits higher,
and wiring in an external verifier. Agreement among agents is not evidence that they are
right: without an independent channel to the answer, debate redistributes the uncertainty
the agents already have rather than reducing it.

\section{Limitations}
\label{sec:limitations}

The reduction is an identity, so the wall's strength is empirical and the measured
AUROCs are lower bounds: the robust claim is that separability is well below $1$. The
steering result is on ARC for the two adequately-powered families, and the
strong-$\alpha$ asymmetry on Qwen-3B is not a usable brake because it does not spare
beneficial revision (Appendix~\ref{app:steering}). The society runs are small and
mostly same-family, on multiple-choice items. The mean-field model is illustrative,
and the warrant arm needs human validation.

Several axes are left to future work; the reduction predicts each stays bounded
unless it adds an independent channel to the answer. These include deploy-time
\emph{utility} curves under asymmetric harmful/beneficial costs (beyond AUROC);
open-ended and tool- or retrieval-augmented settings and alternative debate
protocols (rounds, aggregation rules); broader cross-provider and base-vs-instruct
heterogeneity in the society experiments; further post-hoc signals such as
rationale--answer consistency \citep{fan2025peerguard} or text classifiers on
revision rationales; and adaptive or token-wise steering
\citep{wang2024act,nguyen2025matsteer}, which the reduction subsumes as further
functions of deploy-time signals.

\section{Conclusion}
\label{sec:conclusion}

Filtering harmful peer revisions while keeping beneficial ones is the same task as
judging one's own correctness, and is therefore limited by self-knowledge.
Empirically that ceiling is well below $1$ and is reached but not exceeded by every
deploy-time signal we test, including causal white-box steering. At population scale
this limit combines with correlated errors and majority revision into a
difficulty-gated, confident-wrong cliff that the obvious rescues do not prevent. The
reliable way to lower correctness-uncertainty is to add information the system does
not already hold: better-calibrated self-knowledge, or an independent, more capable
verifier, not a post-hoc filter on the conversation.

\newpage
\section*{Acknowledgments}
This work used Jetstream2 at Indiana University through ACCESS allocation CIS260254 from
the Advanced Cyberinfrastructure Coordination Ecosystem: Services \& Support (ACCESS)
program, which is supported by U.S. National Science Foundation grants \#2138259,
\#2138286, \#2138307, \#2137603, and \#2138296. Results were also obtained using the
Chameleon testbed, supported by the National Science Foundation. We thank the Jetstream2,
ACCESS, and Chameleon support teams for the computational infrastructure used in this work.

\bibliographystyle{iclr2026_conference}
\bibliography{refs}

\appendix

\clearpage

\section{The reduction: selection slack and causal interventions}
\label{app:proof}
Proposition~\ref{prop:reduction} gives the on-flip identity. Two refinements.
\emph{(i) Selection slack.} The conditional ceiling
$\mathrm{AUROC}^{*}_{\textit{correct}_0}(S\mid F)$ equals the global self-knowledge
$\mathrm{AUROC}^{*}_{\textit{correct}_0}(S)$ iff
$\text{flip}\perp\textit{correct}_0\mid S$; the empirically small difference is the
only gap. \emph{(ii) Causal interventions.} For any post-processing $S'=T(S,U)$ with
$U\perp(\textit{correct}_0,V)$ (including activation steering $a\mapsto a+\alpha d$
along the diff-in-means correctness direction), the chain
$\textit{correct}_0\to S\to S'$ is Markov, so by the data-processing inequality
\citep{cover2006elements} $\mathrm{AUROC}^{*}_V(S'\mid F)\le
\mathrm{AUROC}^{*}_{\textit{correct}_0}(S\mid F)$. We treat this as motivation only,
since steering also changes \emph{which} items flip; the white-box case is settled
empirically in Appendix~\ref{app:steering}.

\section{Self-knowledge ceiling: per-cell results}
\label{app:ceiling}
Table~\ref{tab:ceiling-cells} gives the per-cell numbers behind Figure~\ref{fig:ceiling}
(local 4-bit families): self-knowledge AUROC ($x$), harmful-vs-beneficial separation
($y$), and flips per cell. Seven of eight cells have $y\le x$; the exception
(gemma/ARC) has $35$ flips and a bootstrap gap CI spanning $0$. The structural checks
(held-out families, scale ladder, reasoning models) are summarized in
Table~\ref{tab:ceiling} in the main text.

\begin{table}[h]
\centering\small
\caption{Per-cell ceiling, local 4-bit families. $x=$ self-knowledge AUROC,
$y=$ harmful-vs-beneficial separation on flips.}
\label{tab:ceiling-cells}
\begin{tabular}{llccc}
\toprule
Model & Dataset & $x$ (self-know.) & $y$ (separation) & $n_{\text{flip}}$ \\
\midrule
Qwen2.5-7B   & ARC        & $0.89$ & $0.86$ & $143$ \\
Qwen2.5-7B   & TruthfulQA & $0.68$ & $0.66$ & $130$ \\
Llama-3.1-8B & ARC        & $0.75$ & $0.74$ & $124$ \\
Llama-3.1-8B & TruthfulQA & $0.66$ & $0.57$ & $111$ \\
Mistral-7B   & ARC        & $0.67$ & $0.67$ & $78$  \\
Mistral-7B   & TruthfulQA & $0.64$ & $0.61$ & $79$  \\
gemma-2-9B   & ARC        & $0.72$ & $0.86$ & $35$  \\
gemma-2-9B   & TruthfulQA & $0.71$ & $0.68$ & $53$  \\
\bottomrule
\end{tabular}
\end{table}

\paragraph{Estimating the ceiling.} For each signal $S$ we estimate
$\mathrm{AUROC}^{*}_{\textit{correct}_0}(S\mid F)$ by fitting the strongest scoring
rule $g$ we can for that signal, logistic/threshold scores for scalar signals
(verbalized confidence, logprob margin, entropy) and cross-validated linear and
non-linear probes (including the engineered late-layer delta) for activations, and
reporting its AUROC. These are \emph{lower bounds} on the supremum over all $g$; the
claim is therefore that separability stays well below $1$ even for the strongest
scorer we found, not that we compute the exact Bayes ceiling.

The behavioral gate catches harmful revisions at $95.3\pm1.4\%$ ($\alpha{=}0.05$,
25 seeds) but its advantage over plain confidence is $-0.4\pm3.0$\,pp, not better.
Beyond AUROC, the operating curve makes the cost explicit (Figure~\ref{fig:gate}):
across all thresholds the engineered feature gate sits on or below plain confidence,
and catching more harmful revisions always preserves fewer beneficial ones. So
whether any threshold yields a net-accuracy gain depends entirely on the
harmful:beneficial base rate, not on selectivity, and the gate never dominates plain
confidence.

The probe rises from a naive single-layer $\approx 0.61$ to $0.81$--$0.83$
with late-layer pressure deltas (permutation null $0.46$--$0.49$); an answer-content
control on the non-flip set gives $0.79$, so the signal is genuine
correctness-awareness, and it does not exceed the $\approx 0.75$--$0.83$
self-knowledge band.

\paragraph{A rationale text classifier (additional baseline).} A richer, text-based
signal does not beat the self-knowledge signals either. Taking the model's
free-text \emph{justification} for each revision (Qwen-7B/ARC; $n{=}452$ flips, $333$
harmful / $119$ beneficial) and training a TF--IDF $+$ logistic classifier with
item-grouped cross-validation to predict harmful vs.\ beneficial gives AUROC $0.46$,
\emph{at chance} (label-permutation null $\approx 0.50$), far below the
self-knowledge ceiling. The model's stated reasoning carries no usable signal about
whether its own revision was harmful, consistent with the reduction and with revision
being content- rather than self-diagnostic.

\paragraph{Fusing signals (additional baseline).} Combining the deploy-time signals
does not clear the wall either (Table~\ref{tab:ensemble}).

\begin{table}[h]
\centering\small
\caption{Fusing deploy-time signals does not clear the wall. Per cell: the best
\emph{single}-signal self-knowledge AUROC (verbalized confidence, logprob confidence,
or entropy) vs.\ a logistic-regression \emph{ensemble} of all three (out-of-fold
$5$-fold CV, $n{=}120$/cell). The ensemble does not beat the best single signal
(mean $\Delta=-0.05$).}
\label{tab:ensemble}
\setlength{\tabcolsep}{6pt}
\begin{tabular}{llccc}
\toprule
Model & Dataset & best single & ensemble (CV) & $\Delta$ \\
\midrule
Qwen-1.5B & ARC        & $0.72$ & $0.67$ & $-0.06$ \\
Qwen-1.5B & TruthfulQA & $0.73$ & $0.79$ & $+0.06$ \\
Qwen-3B   & ARC        & $0.80$ & $0.67$ & $-0.13$ \\
Qwen-3B   & TruthfulQA & $0.85$ & $0.76$ & $-0.09$ \\
\midrule
\multicolumn{2}{l}{mean} & $0.78$ & $0.72$ & $-0.05$ \\
\bottomrule
\end{tabular}
\end{table}

\paragraph{Sampling-based self-consistency (additional baseline).} A signal class
outside the single-forward-pass argument also stays under the ceiling, and does not
improve with more samples (Table~\ref{tab:selfconsistency}).

\begin{table}[h]
\centering\small
\caption{Sampling-based self-consistency does not clear the wall. Per cell we draw
$k$ temperature samples ($T{=}0.7$) and score the answer-distribution spread as a
detector of majority-vote correctness. The best AUROC stays within the single-pass
band ($0.64$--$0.89$) and does not rise as $k$ grows to $24$.}
\label{tab:selfconsistency}
\setlength{\tabcolsep}{6pt}
\begin{tabular}{llccc}
\toprule
Model & Dataset & $k$ & $n$ & best AUROC (95\% CI) \\
\midrule
Llama-3.1-8B & ARC        & $6$ & $300$ & $0.71\ [0.64,0.78]$ \\
Llama-3.1-8B & TruthfulQA & $6$ & $160$ & $0.69\ [0.62,0.77]$ \\
Qwen2.5-7B   & ARC        & $6$ & $299$ & $0.61\ [0.54,0.69]$ \\
Qwen2.5-7B   & TruthfulQA & $6$ & $238$ & $0.57\ [0.53,0.61]$ \\
\midrule
Qwen2.5-14B & TruthfulQA & $6$  & $100$ & $0.53\ [0.50,0.60]$ \\
Qwen2.5-14B & TruthfulQA & $12$ & $100$ & $0.56\ [0.51,0.63]$ \\
Qwen2.5-14B & TruthfulQA & $24$ & $100$ & $0.57\ [0.51,0.66]$ \\
\bottomrule
\end{tabular}
\end{table}

\section{Steering: equivalence test and dose--response grid}
\label{app:steering}
We steer $3$ families $\times\,3$ seeds $\times\,n{=}300$ on ARC (beneficial-arm
$n{=}107$--$320$). A pre-registered two-one-sided-tests (TOST) equivalence test at
$\pm0.05$ on $\Delta H-\Delta B$ uses a conservative independent-binomial CI. TOST
passes for Llama-3.1-8B over $\alpha\in[-2,+2]$ (max $|\Delta H-\Delta B|=0.008$) and
for Qwen2.5-3B over $\alpha\in[-4,+0.5]$ (max $0.006$); for $|\alpha|\le 1$ both
satisfy $|\Delta H-\Delta B|\le 0.03$. Qwen2.5-7B ($n{=}107$) is consistent with the
null but noisier.

At strong positive $\alpha$, Qwen-3B suppresses harmful flips by a larger factor
than beneficial ones (harmful/beneficial fall to $0.76$/$0.87$ of baseline at
$\alpha{=}{+}2$ and $0.35$/$0.69$ at $\alpha{=}{+}4$; baselines $H(0){=}0.96$,
$B(0){=}0.98$), so $\Delta H-\Delta B$ reaches $-0.11$ and $-0.33$
(Table~\ref{tab:steer-cells}). This asymmetry is genuine but is not a usable brake:
each such setting also drops beneficial revision by $13$--$31$\,pp, and Llama-8B
shows no comparable effect (its only large asymmetry, $+0.089$ at $\alpha{=}{-}4$,
has the opposite sign). The harmful and beneficial steering directions have cosine
$+0.585$ (Qwen-3B pilot), consistent with a largely shared axis.

\begin{table}[h]
\centering\small
\caption{Steering dose--response, pooled over 3 seeds. $\Delta H,\Delta B$ are
changes in harmful/beneficial flip rate relative to $\alpha{=}0$.}
\label{tab:steer-cells}
\setlength{\tabcolsep}{5pt}
\begin{tabular}{rcccccc}
\toprule
& \multicolumn{3}{c}{Llama-3.1-8B ($b_n{=}208$)} & \multicolumn{3}{c}{Qwen2.5-3B ($b_n{=}320$)} \\
\cmidrule(lr){2-4}\cmidrule(lr){5-7}
$\alpha$ & $\Delta H$ & $\Delta B$ & $\Delta H{-}\Delta B$ & $\Delta H$ & $\Delta B$ & $\Delta H{-}\Delta B$ \\
\midrule
$-4$   & $-0.094$ & $-0.183$ & $+0.089$ & $+0.017$ & $+0.016$ & $+0.002$ \\
$-2$   & $-0.013$ & $-0.005$ & $-0.008$ & $+0.021$ & $+0.016$ & $+0.005$ \\
$-1$   & $-0.001$ & $+0.000$ & $-0.001$ & $+0.017$ & $+0.013$ & $+0.005$ \\
$-0.5$ & $+0.001$ & $+0.000$ & $+0.001$ & $+0.009$ & $+0.009$ & $-0.001$ \\
$+0.5$ & $+0.003$ & $+0.000$ & $+0.003$ & $-0.019$ & $-0.012$ & $-0.006$ \\
$+1$   & $+0.001$ & $+0.000$ & $+0.001$ & $-0.074$ & $-0.047$ & $-0.027$ \\
$+2$   & $-0.006$ & $+0.000$ & $-0.006$ & $-0.236$ & $-0.128$ & $-0.108$ \\
$+4$   & $-0.075$ & $-0.038$ & $-0.037$ & $-0.631$ & $-0.303$ & $-0.328$ \\
\bottomrule
\end{tabular}
\end{table}

\paragraph{SAE-feature steering (robustness to the direction source).}
The null does not depend on our choice of a single linear contrastive-activation-addition
(CAA) direction \citep{rimsky2023caa}, nor on the
strength of the white-box method. We re-source the steering direction from a sparse-autoencoder feature space \citep{cunningham2024sparse} (the
Goodfire L19 SAE for Llama-3.1-8B), taking the contrastive top-$64$ correctness features,
decoding them back to the residual stream, and rescaling to the CAA direction's magnitude
so that $\alpha$ denotes the same effect size; only the \emph{direction} changes. The
resulting direction is genuinely distinct (cosine $0.31$--$0.54$ to CAA across seeds and
datasets), yet the null is unchanged (Table~\ref{tab:steer-sae}). On ARC, TOST passes over
$\alpha\in[-2,+2]$ (max $|\Delta H-\Delta B|=0.026$); on TruthfulQA, a harder task with a
balanced beneficial arm ($b_n{=}395$ vs.\ $183$ on ARC), which removes the near-ceiling
beneficial baseline of the ARC split, TOST passes over $\alpha\in[-2,+1]$ (max $0.006$),
and for $|\alpha|\le 1$ both datasets satisfy $|\Delta H-\Delta B|\le 0.014$. As with CAA,
the only departures are at extreme $|\alpha|$ and are non-selective co-collapse (e.g.\
ARC $\alpha{=}{-}4$: $\Delta H{=}{-}0.135$, $\Delta B{=}{-}0.038$; TruthfulQA
$\alpha{=}{+}4$: $\Delta H{=}{-}0.166$, $\Delta B{=}{-}0.066$), never a setting that
suppresses harm while sparing benefit. Results pool $3$ seeds at $n{=}300$; the
$k{\in}\{16,256\}$ feature-budget variants reproduce the in-range equivalence.

\begin{table}[h]
\centering\small
\caption{SAE-feature steering (Goodfire L19, Llama-3.1-8B), pooled over 3 seeds,
$\Delta H,\Delta B$ relative to $\alpha{=}0$. A direction sourced from SAE features
rather than the CAA mean-difference gives the same null on both datasets.}
\label{tab:steer-sae}
\setlength{\tabcolsep}{5pt}
\begin{tabular}{rcccccc}
\toprule
& \multicolumn{3}{c}{ARC ($b_n{=}183$)} & \multicolumn{3}{c}{TruthfulQA ($b_n{=}395$)} \\
\cmidrule(lr){2-4}\cmidrule(lr){5-7}
$\alpha$ & $\Delta H$ & $\Delta B$ & $\Delta H{-}\Delta B$ & $\Delta H$ & $\Delta B$ & $\Delta H{-}\Delta B$ \\
\midrule
$-4$   & $-0.135$ & $-0.038$ & $-0.097$ & $-0.079$ & $-0.104$ & $+0.025$ \\
$-2$   & $-0.026$ & $+0.000$ & $-0.026$ & $-0.048$ & $-0.053$ & $+0.006$ \\
$-1$   & $-0.004$ & $+0.000$ & $-0.004$ & $-0.016$ & $-0.020$ & $+0.004$ \\
$-0.5$ & $+0.000$ & $+0.000$ & $+0.000$ & $-0.006$ & $-0.008$ & $+0.002$ \\
$+0.5$ & $+0.000$ & $-0.011$ & $+0.011$ & $-0.006$ & $+0.000$ & $-0.006$ \\
$+1$   & $-0.003$ & $-0.016$ & $+0.014$ & $-0.008$ & $-0.005$ & $-0.003$ \\
$+2$   & $-0.026$ & $-0.027$ & $+0.001$ & $-0.046$ & $-0.023$ & $-0.023$ \\
$+4$   & $-0.077$ & $-0.060$ & $-0.017$ & $-0.166$ & $-0.066$ & $-0.101$ \\
\bottomrule
\end{tabular}
\end{table}

\paragraph{Scale robustness (32B).}
The equivalence is not an artifact of model scale. Steering Qwen2.5-32B on ARC (CAA
direction, $3$ seeds, $n{=}300$; pooled beneficial arm $b_n{=}56$) keeps the harmful
and beneficial arms together in the deploy-relevant range:
$\max_{|\alpha|\le 1}|\Delta H-\Delta B|=0.07$, with accuracy on both arms flat near
$0.80$ throughout (Table~\ref{tab:steer-32b}). As at $\le$8B, the only larger gaps
appear at extreme $|\alpha|\ge 2$ and are non-selective co-collapse, with accuracy
\emph{falling} on both arms. We do not read the equivalence test on the 32B
TruthfulQA cell: there positive steering saturates \emph{both} arms to ceiling
accuracy ($h_{\text{acc}},b_{\text{acc}}\to 1.00$ at $\alpha\ge 1$), which rails the
flip-rate metric and makes the dose--response uninformative: a ``gap'' produced by
driving both arms to fully correct is not a brake that spares one arm.

\begin{table}[h]
\centering\small
\caption{Steering dose--response at 32B (Qwen2.5-32B, ARC), pooled over 3 seeds.
$\Delta H,\Delta B$ are changes in harmful/beneficial flip rate relative to
$\alpha{=}0$; the last two columns give per-arm accuracy. The arms track together in
range ($|\Delta H-\Delta B|\le 0.07$ for $|\alpha|\le1$) with accuracy flat near
$0.80$; departures at $|\alpha|\ge2$ are co-collapse (accuracy falls on both arms).}
\label{tab:steer-32b}
\setlength{\tabcolsep}{5pt}
\begin{tabular}{rccccc}
\toprule
$\alpha$ & $\Delta H$ & $\Delta B$ & $\Delta H{-}\Delta B$ & $h_{\text{acc}}$ & $b_{\text{acc}}$ \\
\midrule
$-4$   & $+0.295$ & $-0.089$ & $+0.384$ & $0.50$ & $0.53$ \\
$-2$   & $+0.015$ & $-0.088$ & $+0.103$ & $0.78$ & $0.77$ \\
$-1$   & $-0.001$ & $-0.073$ & $+0.072$ & $0.80$ & $0.82$ \\
$-0.5$ & $-0.011$ & $-0.034$ & $+0.024$ & $0.81$ & $0.86$ \\
$+0.5$ & $+0.006$ & $-0.017$ & $+0.023$ & $0.79$ & $0.88$ \\
$+1$   & $+0.020$ & $-0.033$ & $+0.053$ & $0.78$ & $0.86$ \\
$+2$   & $+0.072$ & $-0.033$ & $+0.105$ & $0.73$ & $0.86$ \\
$+4$   & $+0.207$ & $-0.143$ & $+0.350$ & $0.59$ & $0.73$ \\
\bottomrule
\end{tabular}
\end{table}

\section{Transfer beyond multiple choice: GSM8K and MATH}
\label{app:transfer}
We rerun the self-knowledge protocol on two exact-match-checkable free-response tasks:
GSM8K (integer answers) and MATH (boxed expressions, normalized before comparison).
Two models (Qwen2.5-7B, Llama-3.1-8B), greedy decoding, $n{=}200$ per cell.
Self-knowledge AUROC (best deploy-time signal; bootstrap $95\%$ CI on GSM8K): on GSM8K, Qwen $0.85$ $[0.77,0.93]$
and Llama $0.69$ $[0.56,0.80]$; on MATH, Qwen $0.58$ and Llama $0.62$. Self-knowledge
stays imperfect ($<1$) throughout, the same regime as the multiple-choice ceiling, so
the wall is not specific to the multiple-choice format.

\section{Population: dynamics and the mean-field model}
\label{app:robust}
\label{app:meanfield}
The basin numbers (\S\ref{sec:cliff}) pool $8$ TruthfulQA runs and $8$ ARC runs; the
per-condition rescue results are in Table~\ref{tab:robust} (main text). The minority
sets are small ($n\approx 5$--$21$ qualifying items per run), which we read with that
caveat. The heterogeneous diversity rescue is replicated over three seeds on both
datasets, with the society below its best member in every seed: TruthfulQA (14B+OLMo+Phi)
society $0.60$--$0.66$ vs.\ best $0.71$--$0.77$, and ARC (7B+8B+3B) society
$0.80$--$0.87$ vs.\ best $0.86$--$0.90$.

\paragraph{Mean-field model.} Consider $N\to\infty$ agents adopting the
confidence-weighted local-majority answer. Let $\mu$ be the error-correlation
(fraction of wrong agents on one shared distractor) and $r=w_d/w_c$ the
confidence-ratio of wrong to correct agents. The accuracy map has an unstable
interior fixed point (the basin boundary; stated as Proposition~\ref{thm:meanfield})
\begin{equation}
p^{*}=\frac{\mu r}{1+\mu r}, \qquad G'(p^{*})>1,
\label{eq:pstar}
\end{equation}
so the population is bistable with separatrix $p^{*}$. At $r{=}1$ this is the
calibration-blind DeGroot/Golub--Jackson regime; the calibration axis $r$ is the
addition. We estimate $\mu$ as the fraction of wrong agents concentrated on the
single most common distractor (per item, averaged) and $r$ as the ratio of mean
round-1 confidence of wrong to correct agents; both are read from the round-1
kernels and are \emph{not} fit to the society outcomes. From these,
$r\approx 0.91$--$1.05$ (mean $0.95$); averaging, Eq.~\eqref{eq:pstar} ($\mu{=}0.61$)
gives a measured separatrix $p^{*}\approx 0.368$ and a calibrated counterfactual
$\approx 0.315$ (gap $\approx +0.05$), and the simulation reproduces both with no
refit ($0.350$/$0.283$, gap $+0.07$). Because showing vs.\ hiding peer confidence
does not change the live outcome (\S\ref{sec:cliff-mechanism}), $r$ enters only as a
structural property, not an operator signal; we therefore read it as a reduced-form
stickiness asymmetry and treat the empirical transition as the result.

\begin{proposition}[Bistability]
\label{thm:meanfield}
Under the kernel above, $G$ has the unstable interior fixed point
$p^{*}=\mu r/(1+\mu r)$ with $G'(p^{*})>1$; the dynamics are bistable with separatrix
$p^{*}$, undergoing a transcritical bifurcation as $\mu r\to 0^{+}$.
\end{proposition}

\begin{figure}[t]
\centering
\includegraphics[width=\linewidth]{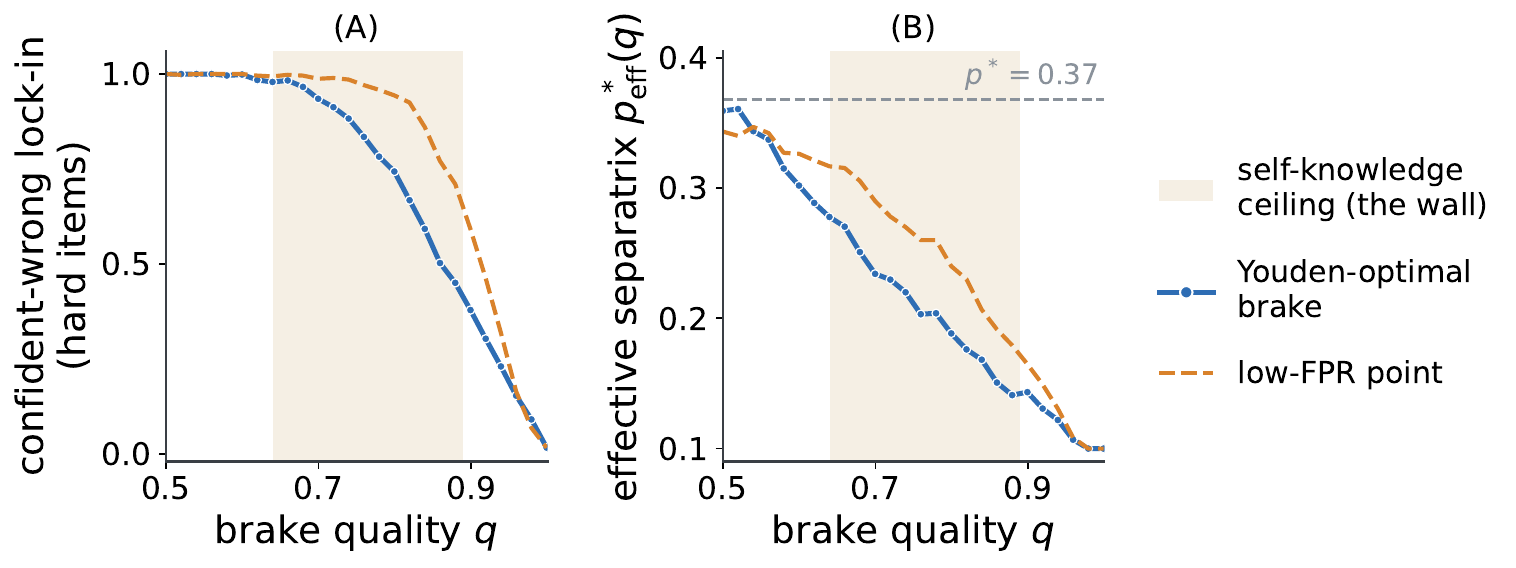}
\caption{The wall bounds the brake. A hypothetical selective brake of quality $q$
(correctness-detector AUROC) blocks harmful revisions at its Youden-optimal point.
\textbf{(A)} confident-wrong lock-in on hard items (groups $\ge$4:1 wrong) vs.\ $q$;
\textbf{(B)} effective separatrix $p^{*}_{\text{eff}}(q)$. The measured ceiling
($q\in[0.64,0.89]$, shaded) barely dents the cliff; only $q\gtrsim0.96$ averts it
(dashed: a low-FPR variant).}
\label{fig:brake}
\end{figure}

\paragraph{The wall bounds the brake.} We extend the dynamics above with a hypothetical
selective brake: a correctness detector of quality $q=\mathrm{AUROC}$ blocks harmful
revisions at its Youden-optimal operating point (binormal equal-variance ROC,
true/false positive rate $\mathrm{TPR}(q)=\Phi(\Phi^{-1}(q)/\sqrt2)$, $\mathrm{FPR}=1-\mathrm{TPR}$; $q{=}0.5$
recovers the un-braked cliff, $q{=}1$ is an ideal brake) and sweep $q$
(Figure~\ref{fig:brake}). On the hardest items (groups starting $\ge$4:1 wrong),
confident-wrong lock-in falls from $1.00$ (no brake) to $0.98$ at $q{=}0.64$ and $0.41$
at $q{=}0.89$, reaching $\le0.05$ only at $q\gtrsim0.96$; the effective separatrix moves
$0.36\to0.28\to0.14$ over the same range. A brake anywhere in the measured
self-knowledge band ($q\in[0.64,0.89]$) leaves the cliff largely intact, so averting it
requires a selectivity the wall rules out.

\section{Warrant arm (preliminary)}
\label{app:warrant}
Single-seed, single-model (Qwen-7B), ARC, same-family self-judge. A neutral,
non-priming warrant flags $\approx 37\%$ of harmful revisions ($n{=}76$) as
violating the model's own pre-commitment, which keyword and SOCIAL/EVIDENCE judges
miss. A change-priming phrasing inflates the flip rate ($31.4\%\to40.2\%$); the neutral
phrasing leaves the flip rate at baseline ($29.1\%$), so the warrant exposes rather than
suppresses. This diagnostic is preliminary and judge-dependent; human adjudication
is needed before treating the absolute rate as stable.

\section{Experimental details}
\label{app:details}

\paragraph{Checkpoints and data.} Local revision/probe/steering experiments use
4-bit (MLX) builds of Qwen2.5-7B-Instruct, Qwen2.5-3B-Instruct,
Meta-Llama-3.1-8B-Instruct, Mistral-7B-Instruct-v0.3, and gemma-2-9b-it. Held-out
checks use \texttt{allenai/OLMo-2-1124-7B-Instruct} and
\texttt{microsoft/Phi-3.5-mini-instruct} at bf16 (Transformers); the scale ladder
uses \texttt{Qwen/Qwen2.5-\{1.5B,7B,14B,32B,72B\}-Instruct} (72B at bf16, sharded over two A100-80GB, $n{=}200$ per cell $\times$ 3 seeds; ARC base accuracy $0.94$--$0.955$ leaves few flips, so the reported quantity is \texttt{selfknow\_global}); reasoning uses
\texttt{deepseek-ai/DeepSeek-R1-Distill-Qwen-7B} and \texttt{-Llama-8B}. Data are
\texttt{allenai/ai2\_arc} (ARC-Challenge, \texttt{test}) and
\texttt{truthfulqa/truthful\_qa} (\texttt{multiple\_choice}, \texttt{mc1},
\texttt{validation}), seed-subsampled.

\paragraph{Decoding.} Correctness, probe, and steering reads are deterministic: the
answer label is scored from the next-token logits after the JSON prefix
\verb|{"judgment":"(| (\texttt{do\_sample=False}), so all AUROCs use logprobs, not
sampling. Society runs sample each agent at temperature $0.7$ (max $40$ tokens/turn),
$N$ agents, a round~0 plus three revision rounds, shown/hidden peer-confidence
conditions sharing round~0; the basin gap uses an item-level bootstrap $95\%$ CI.
The conformal gate is class-conditional split-conformal ($\alpha{=}0.05$, 25 seeds);
the probe uses item-grouped CV with a 20-shuffle permutation null.

\paragraph{Prompt templates.} gemma has no system role and receives the system text
prepended to the first user turn.

{\small
\begin{verbatim}
[system] You are a careful assistant solving multiple-choice questions.
         Follow the requested JSON schema exactly.

[round 1, user]
Answer the following multiple-choice question.
<JSON INSTRUCTIONS>
Question:
{question}
Options:
{(A) ... (B) ... (C) ... (D) ...}

[round 2, user]  (peer turn; round-1 answer echoed as assistant)
Other people have now shared their initial answers:
{peer summary}
Answer the same multiple-choice question again.
<JSON INSTRUCTIONS>

<JSON INSTRUCTIONS> =
Return valid JSON only, with exactly this format:
{"judgment":"(A)","confidence":7}
Requirements:
- "judgment" must be exactly one option label shown, like "(A)".
- Return the choice label only, not the option text.
- "confidence" must be an integer from 1 to 10 on this scale:
  1 = guessing randomly; 3 = slight inclination, quite uncertain;
  5 = think it is right but could easily be wrong; 7 = fairly
  confident with some doubt; 9 = very confident; 10 = virtually certain.
- Do not include reasoning, markdown, or text outside the JSON object.

{peer summary} shown:  "<k> chose (X) (avg confidence <c>/10)"
{peer summary} hidden: "<k> chose (X)"
\end{verbatim}
}

\end{document}